\documentclass[sigconf]{acmart}
\renewcommand\footnotetextcopyrightpermission[1]{}
\usepackage{multirow}
\usepackage{hyperref}

\AtBeginDocument{%
  }

\setcopyright{acmlicensed}
\copyrightyear{2025} 
\acmYear{2025} 
\setcopyright{acmlicensed}\acmConference[MM '25]{Proceedings of the 33rd
	ACM International Conference on Multimedia}{October 27--31, 2025}{Dublin,
	Ireland}
\acmBooktitle{Proceedings of the 33rd ACM International Conference on
	Multimedia (MM '25), October 27--31, 2025, Dublin, Ireland}
\acmDOI{10.1145/3746027.3755182}
\acmISBN{979-8-4007-2035-2/2025/10}

\begin{document}

\title{Infrared and Visible Image Fusion with Language-Driven Loss in CLIP Embedding Space}


\author{Yuhao Wang}
\affiliation{
	\institution{School of Automation\\Beijing Institute of Technology}
	\city{Beijing}
	\country{China}}
\email{wayhao@bit.edu.cn}
\orcid{0009-0001-5633-7927}

\author{Lingjuan Miao}
\affiliation{
	\institution{School of Automation\\Beijing Institute of Technology}
	\city{Beijing}
	\country{China}}
\email{miaolingjuan@bit.edu.cn}
\orcid{0000-0003-1782-4535}

\author{Zhiqiang Zhou}
\authornote{Corresponding author}
\affiliation{
	\institution{School of Automation\\Beijing Institute of Technology}
	\city{Beijing}
	\country{China}}
\email{zhzhzhou@bit.edu.cn}
\orcid{0000-0001-6871-8236}

\author{Lei Zhang}
\affiliation{
	\institution{School of Automation\\Beijing Institute of Technology}
	\city{Beijing}
	\country{China}}
\email{zhangleiii@bit.edu.cn}
\orcid{0009-0008-5858-6899}

\author{Yajun Qiao}
\affiliation{
	\institution{School of Automation\\Beijing Institute of Technology}
	\city{Beijing}
	\country{China}}
\email{yajun@bit.edu.cn}
\orcid{0009-0007-8120-9146}


\renewcommand{\shortauthors}{Yuhao Wang, Lingjuan Miao, Zhiqiang Zhou, Lei Zhang, and Yajun Qiao}

\begin{abstract}
	Infrared-visible image fusion (IVIF) has attracted much attention owing to the highly-complementary properties of the two image modalities. Due to the lack of ground-truth fused images, the fusion output of current deep-learning based methods heavily depends on the loss functions defined mathematically. As it is hard to well mathematically define the fused image without ground truth, the performance of existing fusion methods is limited. In this paper, we propose to use natural language to express the objective of IVIF, which can avoid the explicit mathematical modeling of fusion output in current losses, and make full use of the advantage of language expression to improve the fusion performance. For this purpose, we present a comprehensive language-expressed fusion objective, and encode relevant texts into the multi-modal embedding space using CLIP. A language-driven fusion model is then constructed in the embedding space, by establishing the relationship among the embedded vectors representing the fusion objective and input image modalities. Finally, a language-driven loss is derived to make the actual IVIF aligned with the embedded language-driven fusion model via supervised training. Experiments show that our method can obtain much better fusion results than existing techniques. The code is available at \textcolor{blue}{\textit{\url{https://github.com/wyhlaowang/LDFusion}}}.
\end{abstract}

\ccsdesc[500]{Computing methodologies~Computer vision}

\keywords{Language-driven fusion loss, Language-expressed fusion objective, Infrared and visible image fusion, Pre-trained vision-language model}

\begin{teaserfigure}
	\centering
	\includegraphics[width=\textwidth]{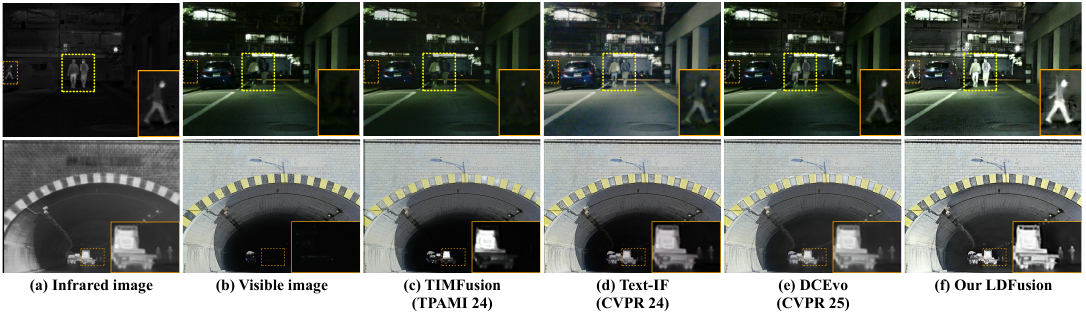} 
	\caption{A comparison of fusion results obtained by our LDFusion and some other state-of-the-art fusion methods.}
	\label{fig:head_results}  
\end{teaserfigure}

\maketitle

\section{Introduction}
Multi-modal image fusion techniques, particularly the ones for infrared and visible image fusion (IVIF), have received considerable attention in recent years\cite{zhang2023visible}. Infrared images reflect thermal radiation information, unaffected by lighting conditions, but often lack sufficient details. Conversely, visible images are often rich in details but sensitive to illumination, smoke and adverse weather. IVIF task aims to generate a fused image that integrate the complementary information from the two modalities. Currently, the deep-learning based methods dominate the field of IVIF. However, since there is no physical ground-truth fused image, the fusion output is greatly dependent on the loss functions defined mathematically for modelling the fused image. 

Researchers have proposed to use various loss functions to supervise fusion network training, including SSIM loss\cite{li2018densefuse,li2023lrrnet}, adversarial loss\cite{rao2023gan,ma2020ddcgan}, contrastive loss\cite{zhu2022clf}, high-level task supervision loss\cite{xie2023semantics,tang2023rethinking}, etc. To improve the fusion results, current deep learning-based methods tend to combine various loss functions. Consequently, the overall loss function becomes increasingly complex. However, without knowing the real fused image, it is difficult for us to really well model the fused output mathematically with analytical loss functions. As a result, the performance of these fusion methods is limited.

\begin{figure}[t]
	\centering
	\includegraphics[width=8.5cm]{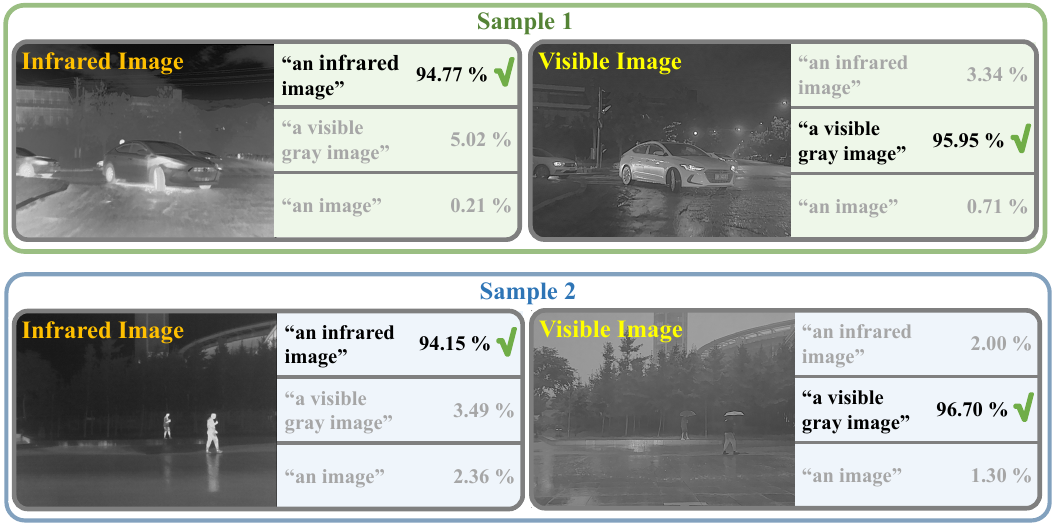} 
	\caption{CLIP can perceive both visible and infrared images.}
	\label{clip_aware}
\end{figure}

Although it is hard to well model the fused image in mathematics, it is possible for us to seek for an alternative solution, \textit{i.e.}, using natural language to express the fusion objective. As we known, natural language can be used more easily to express the concept and intuitive ideas, while it would be much harder to convert them using precise mathematical symbols and formulations. Usually, before we conduct mathematical modeling for a problem, we need to be able to roughly describe the model through language, and then further transform it into mathematical formulas. Mathematical modeling, in contrast, is often much harder than language description, not to mention that many problems can not be explicitly modeled. Based on this principle, to solve the problem of IVIF, this paper attempts to model the fusion objective based on natural language, and proposes a language-driven fusion loss to avoid complex and explicit mathematical modeling. We found that this approach is not only simpler, but also can achieve the fusion results much better than existing methods (see the examples in Fig. \ref{fig:head_results}).

More specifically, to achieve the above goal, we try to leverage the pre-trained vision-language model CLIP\cite{radford2021learning}. As shown in Fig. \ref{clip_aware}, we found that CLIP is able to perceive the infrared images, indicating that it can well encode the infrared image information (as that for the typical visible image) into the multi-modal embedding space of CLIP, in which the vision-language alignment has been achieved. To solve the issue of IVIF, we next establish an embedded language-driven fusion model using the text encoder of CLIP, for which we developed a language-expressed fusion objective, and use the prompts “an infrared image” and “a visible gray image” to describe the input infrared and visible images, respectively. Our goal is then become to make the fusion of infrared and visible images aligned with the language-driven fusion model in CLIP embedding space through supervised training.

The main contributions are summarized as follows: 
\begin{enumerate}
	\item We propose to use nature language to express the whole objective of IVIF, avoiding complex mathematical modeling in current losses in the absence of physical ground truths while obtaining much better fusion results.
	\item A language-driven fusion model is derived in CLIP embedding space, based on which we develop a simple yet highly effective language-driven loss for IVIF. Particularly, by introducing a novel regularization and artifact removal approach, we ensure the trained model works well with high robustness and generalization in practice.
	\item Experiments show a great improvement of fusion quality achieved by the proposed method, revealing the superiority of language in modeling of the fusion output and the potential of pre-trained vision-language model in improving the IVIF performance. 
\end{enumerate}

\section{Related Work}
\textbf{Deep Learning-Based Fusion Methods.} 
Deep learning-based methods have become the mainstream approach to address the inherent challenges of the IVIF task \cite{zhang2021image,zhang2023visible}, including autoencoder (AE)-based methods, convolutional neural network (CNN)-based methods, and semantic-guided fusion methods. Typically, AE-based methods \cite{li2018densefuse, zhao2021efficient, xu2022cufd} were first trained to reconstruct the input images, after which the trained model generates fused images with the help of fusion rules. However, AE-based methods always adopt simple fusion rules (\textit{e.g.}, weighted average, max selection), which lack adaptability and may cause information loss or redundancy. In \cite{xu2020u2fusion}, an end-to-end CNN-based fusion network was proposed for general image fusion. The recently proposed methods \cite{tang2022piafusion,wang2023interactively} tend to use more sophisticated networks to improve the fusion performance. However, these methods rely heavily on various manual designs, making them rather complex, and usually can not achieve satisfactory fusion results in the cases that were not considered. More recently, researchers tend to integrate deep semantic features into the fusion process. One way is to incorporate textual features into the image fusion process. For example, Text-IF \cite{yi2024text} was proposed to integrate textual semantic information into the fusion network. However, these semantic features were extracted from the textual descriptions provided in advance, making it require the manual input of descriptive texts during inference. Another category of methods \cite{liu2024task,liu2024semantic} leverages high-level tasks (\textit{e.g.}, object detection or semantic segmentation) to provide semantic information for the fusion process, but they are constrained by the limited number of object or scene labels, which prevents comprehensive improvements in fusion performance.

\textbf{Loss Functions for IVIF.} 
Due to the lack of physical ground-truth fused image, the fusion result is highly dependent on the used loss functions. Approaches like CDDFuse \cite{zhao2023cddfuse}, ReCoNet \cite{huang2022reconet} and NestFuse \cite{li2020nestfuse} mainly utilized the pixel intensity loss and structure similarity \cite{ma2015perceptual} loss to ensure that fusion results preserve the source image information. To obtain more detailed loss information, perceptual loss \cite{long2021rxdnfuse} was introduced for training. In addition, GAN-based methods \cite{rao2023gan,ma2020ddcgan,xiao2024spdfusion} used adversarial loss to preserve the salient thermal targets and texture details. Nevertheless, GAN-based methods often suffer from unstable training, which impacts the consistency and overall quality of the generated fusion results. More recent studies \cite{liu2022target,zhao2023metafusion,xie2023semantics} utilized high-level tasks to boost performance on downstream vision tasks. However, these methods may not effectively enhance the quality of details in background. Despite the increasingly complex loss functions employed in current IVIF methods, the fusion performance is still not satisfactory.

\section{Proposed Method}
\subsection{The Framework of Our Method}
\begin{figure}[t]
	\centering
	\includegraphics[width=6cm]{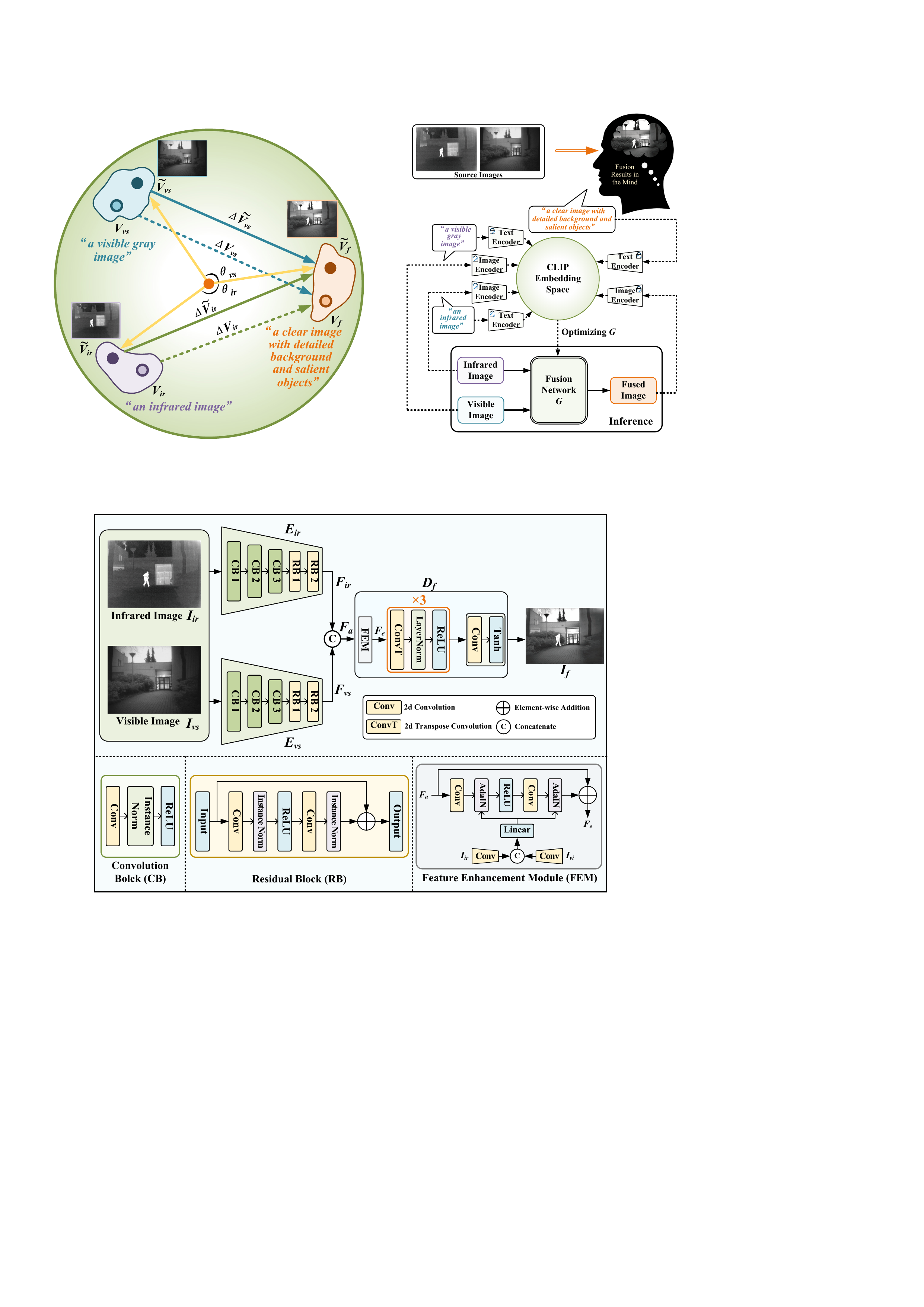} 
	\caption{The Framework of the proposed method. The dashed line represents the language-driven training process, while the solid box denotes the inference process.}
	\label{stage}
\end{figure}
Fig. \ref{stage} illustrates an overview of our method, which includes the language-driven training process and the inference process. In the language-driven training process, the infrared and visible images are input respectively to the image encoder of CLIP to obtain their embedded vectors in CLIP embedding space. At the same time, the texts ``an infrared image'' and ``a visible gray image'' are encoded with the text encoder of CLIP to obtain their corresponding embedded vectors. Besides, we use another text to express the fusion result conceived in our mind and also produce its embedded vector with the text encoder. This enable us to construct an embedded language-driven fusion model in CLIP embedding space, by establishing the relationship between the embedded vector from the text to express fusion objective and the embedded vectors from the texts to describe the different image modalities (will be detailed in Sec. \ref{sec_ldl}). 

Note that in the training process, the fused image obtained by the fusion network $G$ is also input to produce its embedded vector in CLIP embedding space. The language-driven training of the fusion network $G$ is then become to align the relationship between the embedded vectors of input images and fused image with the fusion relationship established by the language-driven model. As the language-driven fusion model is not involved in the inference process, the actual fusion of images does not need to use CLIP, and thus no extra computation is introduced.

\subsection{Language-Expressed Fusion Objective}
The CLIP model that can perceive both infrared and visible image information allows us to express the fusion objective using language, and then encode it into CLIP embedding space. As illustrated in Fig. \ref{stage}, the language expression stems from our mind imaging what the fusion result should be. Using the language, it is possible for us to give a comprehensive description of the fusion effect with a short sentence, \textit{e.g.}, in this paper we describe it as ``a clear image with detailed background and salient objects''. This expression can lead to produce the fused image with pretty rich and comprehensive information, covering both the background and various objects in the image. In contrast, it is difficult to use analytical loss functions to formulate this objective, because we can not define the ``background'' and ``objects'' with mathematical symbols, while the pre-trained language-vision model might have this cognition for an image. Some work \cite{liu2022target,tang2023rethinking} leveraged the object detection or semantic segmentation results to constructed the supervision losses for IVIF. However, these approaches are constrained by a limited number of object or scene labels, whereas CLIP has a generic perception of various objects and backgrounds in reality, not to mention that it can react to the words “salient” and “detailed” in the objective description.

\subsection{Language Prompts for Source Images}
\label{prompt_source}
We use the language prompts ``an infrared image'' and ``a visible gray image'' to describe infrared and visible images respectively, as these prompts align well with source images in CLIP embedding space (\textit{i.e.}, they achieve sufficiently high similarity scores as illustrated in Fig. \ref{clip_aware}). Note that different from existing text-based fusion methods \cite{cheng2023textfusion, yi2024text, zhao2024image}, the purpose of our prompts is not to provide a detailed description of content for each source image, and then use the texts to enhance the image feature extraction and fusion. Instead, we need only to use a set of concise and summary texts (\textit{i.e.}, ``an infrared image'' and ``a visible gray image'') to indicate different source images. We found that this solution would obtain sufficiently good results in practice, as these texts can be always kept well aligned with the source images as illustrated in Fig. \ref{clip_aware}. We have also experimented with alternative prompts, including the specific text description of image content (will be detailed in Sec. \ref{ab_source}). However, our finding is that using the detailed content texts is not necessary, not only because it is time-consuming to provide content descriptions for individual images manually or via AI applications, but also due to the fact that it would harm the generalization of trained fusion model to some extent when the training and testing datasets are not consistent. The underlying insight regarding this issue will be detailed in Sec. \ref{ab_source}.  

\subsection{Language-driven Fusion Model and Loss}
\label{sec_ldl}

\begin{figure}[t]
	\centering
	\includegraphics[width=6cm]{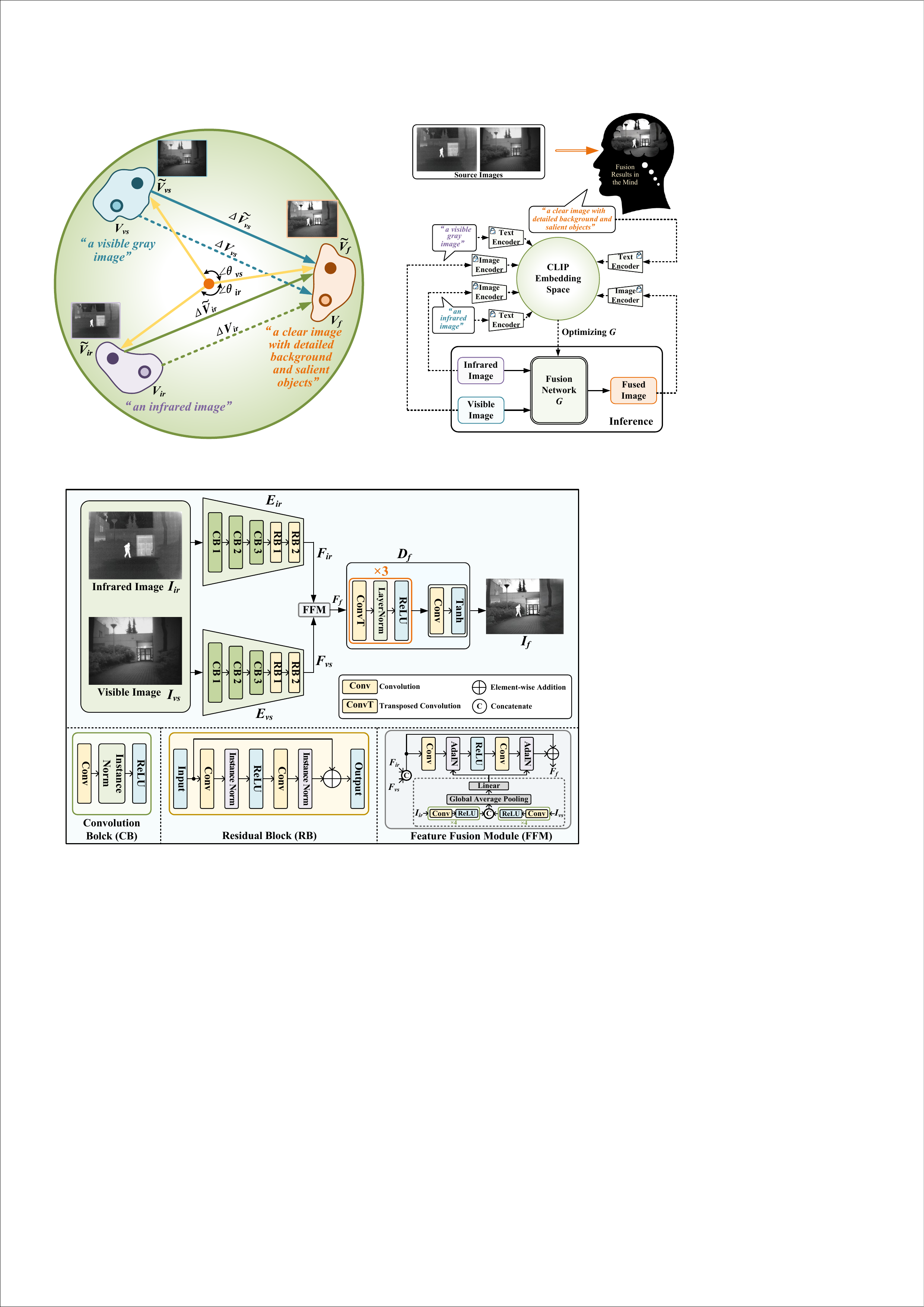} 
	\caption{The schematic of the fusion process in CLIP embedding space. $\Delta V_{vs}$ and $\Delta V_{ir}$ jointly define the transition relationship established by the language-driven fusion model, while $\Delta\tilde{V}_{vs}$ and $\Delta\tilde{V}_{ir}$ denote the actual image fusion relationship.}
	\label{mes}
\end{figure}

\subsubsection{Embedded Language-Driven Fusion Model} 
After having the language-expressed fusion objective encoded in CLIP embedding space, the next issue is to establish a language-driven fusion model, which can guide to generate the described target image from the source images. As shown in Fig. \ref{mes}, let $V_f$ denote the embedded vector from the text to express the fusion objective, and $V_{vs}$, $V_{ir}$ denote the embedded vectors from the texts to describe the visible and infrared images, respectively. Inspired by image generators with CLIP \cite{gal2022stylegan, yang2023zero}, we construct the fusion model by establishing the transition relationship from the source vectors $V_{vs}$, $V_{ir}$ to the objective vector $V_f$. This results in the transition vectors 
$\Delta V_{vs} = V_f - V_{vs}$ and $\Delta V_{ir} = V_f - V_{ir}$ as illustrated by the dash arrows in Fig. \ref{mes}. The combination of these two vectors jointly defines the direction to generate the fused result from the input modalities in the embedding space. 

Based on this principle, we next derive a fusion loss ensuring the generated fused images align with the objective expressed in language. It is worth noting that our approach poses basic differences from previous image generators for style transfer based on directional CLIP loss \cite{gal2022stylegan, kwon2022clipstyler}. Firstly, our research deals with multi-modal image data and aims to address the challenges of heterogeneous image fusion. In particular, for the first time we use a concise and generic language description to express the objective of IVIF, which avoids complex mathematical modeling in current losses in the absence of physical ground truths while obtaining much better fusion results. Secondly, ensuring the language-driven fusion model works properly in practice with good generalization and robustness remains especially challenging, which require novel regularization and training techniques as described below.

\subsubsection{Language-Driven Fusion Loss}
~\\
\textbf{Fusion Direction Loss in CLIP Space.} To obtain the fused image as the language described, we make the actual fusion of infrared and visible images aligned with the embedded language-driven fusion model during the training. For this purpose, the source images and the fused image obtained by the fusion network are input to the image encoder of CLIP (see Fig. \ref{stage}), and obtain the image encoded vectors $\tilde{V}_{vs}$, $\tilde{V}_{ir}$ and $\tilde{V}_{f}$, respectively. As illustrated in Fig. \ref{mes}, due to the alignment of vision-language achieved by CLIP, the image-text encoded vector pairs $\tilde{V}_{vs} \sim V_{vs}$, $\tilde{V}_{ir} \sim V_{ir}$ can be considered to occupy nearby locations in the embedding space, respectively. Thus, to make the actual image fusion aligned with the language-driven model, we need only to force the image-encoded transition vectors $\Delta\tilde{V}_{vs} $ ($\Delta\tilde{V}_{vs} = \tilde{V}_{f} - \tilde{V}_{vs}$) and $\Delta\tilde{V}_{ir}$ ($\Delta\tilde{V}_{ir} = \tilde{V}_{f} - \tilde{V}_{ir}$) parallel to the text-encoded transition vectors $\Delta\tilde{V}_{vs}$ and $\Delta\tilde{V}_{ir}$, respectively. In this case, as illustrated by the two solid arrows in Fig. \ref{mes}, it would drive the fusion towards the language-described direction and reach to the objective state in the embedding space. As a result, we define a fusion direction loss as follows:
\begin{align}
	\begin{gathered}
		\mathcal{L}_d = 1 - \frac{1}{2} \cdot (\frac{\Delta\tilde{V}_{vs} \cdot \Delta{V}_{vs}} {\|\Delta\tilde{V}_{vs}\| \cdot \|\Delta{V}_{vs}\|} + \frac{\Delta\tilde{V}_{ir} \cdot \Delta{V}_{ir}} {\|\Delta\tilde{V}_{ir}\| \cdot \| \Delta{V}_{ir}\|}).\\
	\end{gathered}   
\end{align}

The above loss ensures that $\Delta\tilde{V}_{vs}$ and $\Delta\tilde{V}_{ir}$ are parallel to $\Delta{V}_{vs}$ and $\Delta{V}_{ir}$, respectively. In our implementation, $\mathcal{L}_d$ is performed on the entire image and multi-scale patches. Specially, we achieve this by randomly selecting a number of patches with the dynamic sizes from 112×112 to 180×180 in the images, and applying random affine transformations to each selected patch for producing more diverse samples inspired by Kwon \textit{et al.} \cite{kwon2022clipstyler}. In particular, we propose a novel patch filtering approach to address the risk of textual artifact generation induced by the multi-modal nature of the CLIP model, which will be detailed in Sec. \ref{sec_pf}. 

\textbf{Embedded Vector Direction Regularization.}
The naive fusion direction loss can not ensue high robustness and generalization of the fusion model. The fusion model training should prevent the generated fusion output from being biased toward any of the input modalities, so as to maximize the information contained in the fused result. As shown in Fig. \ref{mes}, this can be assumed to maximize $\angle \theta_{vs}$ (the angle between $\tilde{V}_{f}$ and $\tilde{V}_{vs}$) and $\angle \theta_{ir}$ (the angle between $\tilde{V}_{f}$ and $\tilde{V}_{ir}$) simultaneously in the CLIP embedding space. However, since $\angle \theta_{vs}$ and $\angle \theta_{ir}$ are constrained by each other, optimizing them simultaneously requires a trade-off. As a result, we define the direction regularization term as follows:
\begin{align}
	\begin{gathered}
		\Phi = \left|1 - \cos(\angle\theta_{vs}) \right| + \left|1 - \cos(\angle\theta_{ir})\right| \\
		= \left|1 - \frac{\tilde{V}_{f} \cdot \tilde{V}_{vs}} {\|\tilde{V}_{f}\| \cdot \|\tilde{V}_{vs}\|}\right| + \left|1 - \frac{\tilde{V}_{f} \cdot \tilde{V}_{ir}} {\|\tilde{V}_{f}\| \cdot \|\tilde{V}_{ir}\|}\right|.\\
	\end{gathered}   
\end{align}

Minimizing $\Phi$ would suppress the tendency of $\tilde{V}_{f}$ to become parallel to any of the source vectors $\tilde{V}_{vs}$ and $\tilde{V}_{ir}$, and thus conversely make $\angle \theta_{vs}$ and $\angle \theta_{ir}$ simultaneously as large as allowed. We introduce $\Phi$ into $\mathcal{L}_d$, and obtain a language-driven fusion loss as:
\begin{align}
	\label{ldloss}
	\begin{gathered}
		\mathcal{L}_{d}^{\dagger} = \mathcal{L}_{d} + \lambda\Phi,\\
	\end{gathered}   
\end{align}
where $\lambda$ denotes the weight parameter. Finally, we need to introduce a feature-fidelity loss $\mathcal{L}_v$ to suppress undesired false information generated by the language-driven model. It is defined as follows:
\begin{align}
	\label{feature_loss}
	\begin{gathered}
		\mathcal{L}_{v} = \sum_{k}(\|\psi_{f}^k - max(\psi_{ir}^k,\psi_{vs}^k)\|),\\
	\end{gathered}   
\end{align}
where $\psi_{f}^k$, $\psi_{ir}^k$ and $\psi_{vs}^k$ are the feature maps of fused, infrared, and visible images, extracted by the $k$-th convolution layer of the pre-trained VGG-19 network ($k \in \{3,5,10\} $ in this paper). Therefore, our total loss function consists of the language-driven fusion loss and feature-fidelity loss, formulated as:
\begin{align}
	\label{total_loss}
	\begin{gathered}
		\mathcal{L}_{total} = \mathcal{L}_{d}^{\dagger} + \alpha\mathcal{L}_{v},\\
	\end{gathered}   
\end{align}
in which $\alpha$ is the weight parameter.

\subsubsection{Patch Filter-Based Training for Artifact Removal}
\label{sec_pf}
\begin{figure}[t]
	\centering
	\includegraphics[width=8.5cm]{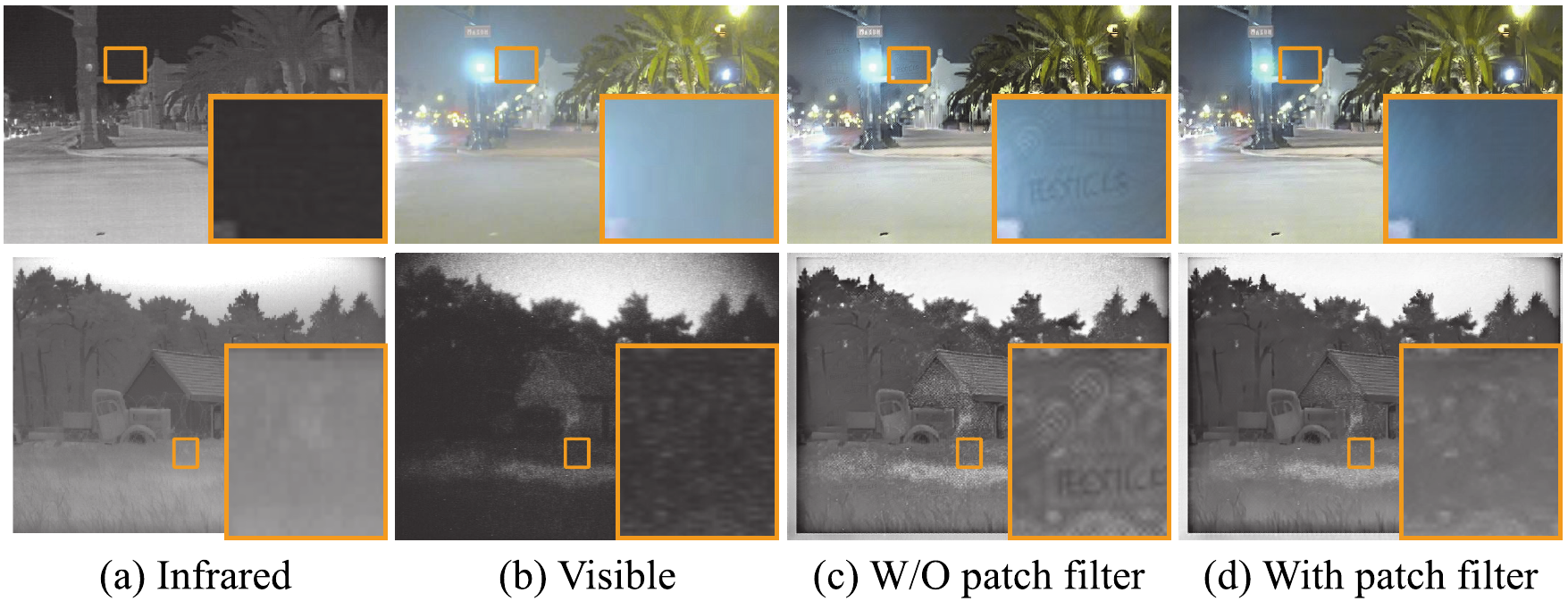} 
	\caption{Some examples of infrared (a) and visible (b) image fusion results without (c) and with (d) the patch filter based on information entropy.}
	\label{pf}
\end{figure}

CLIP-based image generation methods sometimes suffer from the production of textual and other undesirable visual artifacts \cite{kwon2022clipstyler, yang2023zero, xu2024spectralclip}. As shown in Fig. \ref{pf}, the proposed CLIP-based fusion method also creates text-like artifacts (see the lower parts of the enlarged image regions in Fig. \ref{pf}(c)) and other ``imagined'' visual artifacts (see the upper parts of the enlarged regions). While some previous papers \cite{kwon2022clipstyler, yang2023zero} have reported this issue, they seldom addressed it. The issue can be mainly attributed to the multi-modal nature of the CLIP model \cite{radford2021learning}, \textit{i.e.,} CLIP responds consistently to the texts and visual entities that convey the same abstract concepts. As a result, in some cases, it will generate portions of related texts in the image. Besides, CLIP has some ``imagination'' capability, which is probably the cause of creating the ``imagined'' artifacts with spurious content from some vague visual cues (as shown in Fig. \ref{pf}(c)). 

We propose a novel patch filtering approach that is effective yet simple for artifact removal in our fusion method. This approach is proposed based on a key observation: the visual and textual artifacts tend to appear in local regions where the source images have little information. Those regions are relatively flat, mainly with residual noises or image-compression artifacts, which might be used as the visual cues for further image-content generation by CLIP and result in the production of unreal visual entities including the textual artifacts. In contrast, in other regions overwhelmed by valid information, the generation of all the ``imagined'' artifacts can be effectively avoided. As a result, we propose an approach to filter out the patches where the source images are with little information. More specifically, we identify the patches with little information (\textit{e.g.}, the orange boxes in Fig. \ref{pf}(a) and (b)) by evaluating their entropy, which is defined as follows:
\begin{equation}
	H(X) = -\sum_{i=1}^n P(x_i) \log_2 P(x_i)
\end{equation}
where $X$ represents an image patch, $x_i (i=1,2,...,n)$ represent different levels of pixel intensity value, $P(x_i)$ is the probability of intensity $x_i$. We next employ a threshold $\sigma$ (we empirically find that $\sigma=6$ is sufficient to remove artifacts) to filter out the patches with low entropy values, so that they do not contribute to the language-driven fusion loss at region level during the training process. This approach can effectively eliminate the artifacts while maintaining the quality of fusion (as shown in Fig. \ref{pf}(d)). 

\subsection{Fusion Network}
\begin{figure}[h]
	\centering
	\includegraphics[width=6.5cm]{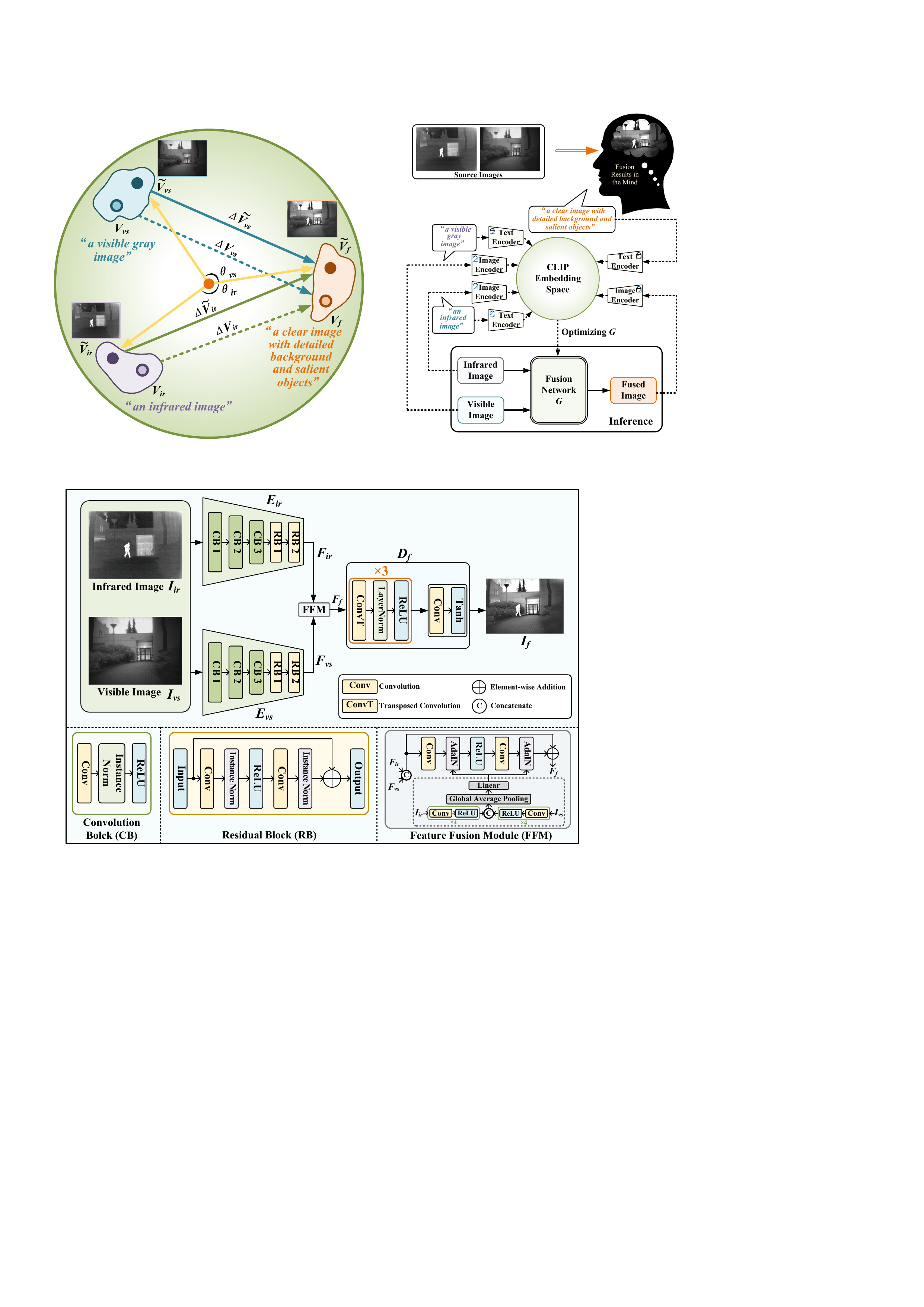} 
	\caption{The structure of the fusion network $G$.}
	\label{network}
\end{figure}

Thanks to the proposed language-driven fusion loss, we can employ a simple fusion network to achieve high-quality image fusion. The structure of our IVIF network is illustrated in Fig. \ref{network}. We use a two-branch encoder ($E_{ir}$ and $E_{vi}$) with the same architecture but different parameters to extract features from each modality. Each branch consists of three convolution blocks (CB) and two residual blocks (RB). The extracted infrared and visible image features ($F_{ir}$ and $F_{vi}$) are then entered into the feature fusion module (FFM), which is constructed by the operator of concatenation and the followed convolution layers. Inspired by \cite{huang2018multimodal} and \cite{gao2022dcdr}, we utilize the adaptive instance normalization (AdaIN) \cite{huang2017arbitrary} instead of the commonly used batch normalization (BN) following the convolution layers, helping to make the distribution of fused features aligned with that of the source image features. For this purpose, we use a convolution module to dynamically generate the AdaIN parameters from source images, which includes a series of convolution with ReLU activation followed by global average pooling and linear layers (as shown in the dashed box of Fig. \ref{network}). The fused features $F_{f}$ are then fed into the decoder $D_{f}$ to reconstruct the final fused image. The decoder consists of three transposed convolution layers with layer normalization and ReLU activation, followed by a convolution layer with Tanh activation to generate the final output.

\section{Experiments}
\begin{figure*}[t]
	\centering
	\includegraphics[width=15cm]{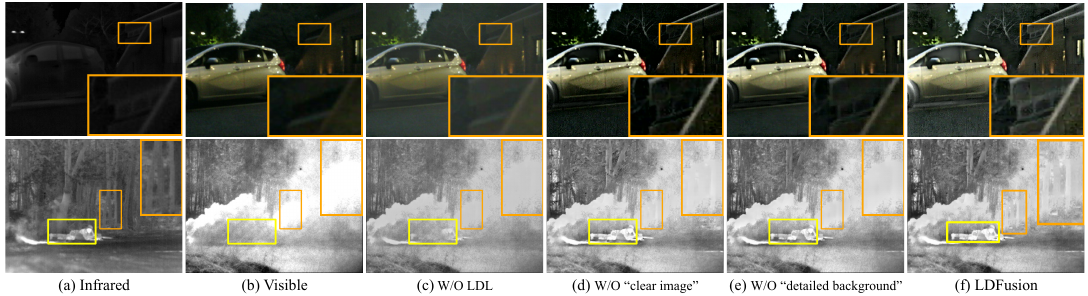} 
	\caption{Some examples of fusion results from ablation study.}
	\label{abf_ld_obj}
\end{figure*}

\subsection{setup}
\textbf{Datasets.} 
We conduct experiments on public datasets M3FD \cite{liu2022target}, RoadScene \cite{xu2020u2fusion}, and TNO \cite{toet2017tno}. Our network was trained on 2940 image pairs from M3FD, and the other 300 pairs are used for testing. In addition, the datasets RoadScene and TNO are utilized for testing as well. For RoadScene, the images are proportionally resized so that their short sides are 768 pixels, due to their original sizes being small and inconsistent. 

\textbf{Compared Methods and Metrics.} We compare our method with 11 state-of-the-art fusion methods, including DCEvo \cite{liu2025dcevo}, Text-IF \cite{yi2024text}, SDCFusion \cite{liu2024semantic}, TIMFusion \cite{liu2024task}, LRRNet \cite{li2023lrrnet}, SwinFusion \cite{ma2022swinfusion} and TarDAL \cite{liu2022target}. Five metrics are used for quantitative evaluation, including information entropy (EN)\cite{roberts2008assessment}, correlation coefficient (CC)\cite{deshmukh2010image}, standard deviation (SD)\cite{ma2019fusiongan}, edge intensity (EI) \cite{zhang2025ddbfusion} and the visual information fidelity for fusion (VIFF)\cite{han2013new}. 

\begin{table}[t]
	\centering
	\renewcommand{\arraystretch}{0.9}
	\caption{Summary of ablation study results on TNO, RoadScene and M3FD. The best results are highlighted in bold.}
	\begin{tabular}{cccccc}
		\hline
		Methods & EN $\uparrow$ & CC $\uparrow$ & SD $\uparrow$ & EI $\uparrow$ & VIFF $\uparrow$ \\
		\hline
		W/O LDL                 & 6.652 & 0.739 & 31.465 & 42.899 & 0.291 \\
		W/O $\Phi$              & 7.172 & 0.754 & 42.579 & 59.466 & 0.588 \\
		W/O CI in LEFO          & 7.322 & 0.752 & 45.906 & 67.424 & 0.582 \\
		W/O DB in LEFO          & 7.185 & 0.550 & 41.209 & 60.878 & 0.550 \\
		LDFusion                & \textbf{7.420} & \textbf{0.758} & \textbf{49.450} & \textbf{69.706} & \textbf{0.728} \\
		\hline
	\end{tabular}
	\label{abt_ldl}
\end{table}

\begin{figure}[t]
	\centering
	\includegraphics[width=8cm]{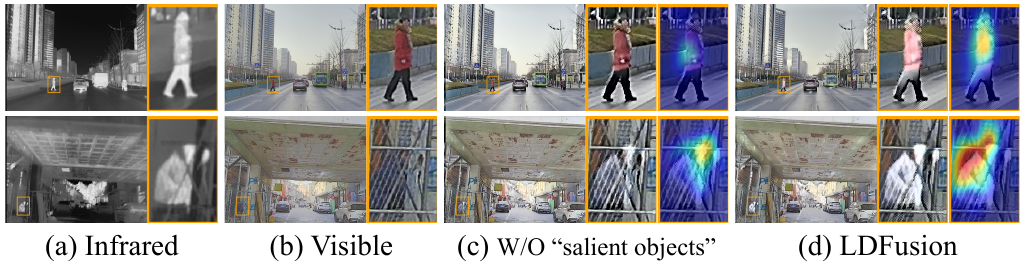} 
	\caption{The examples of fused images with and without ``salient objects''.}
	\label{abf_obj}
\end{figure}

\textbf{Implementation Details.} Our framework is implemented with PyTorch on NVIDIA RTX 3090 GPUs. we employ the Adam optimizer with a learning rate set to 0.001. The training process consists of 70 epochs, with a warm-up phase during the first epoch. After 50 epochs, a decay strategy is implemented, halving the learning rate every 10 epochs. The weight $\alpha$ in Eq. \ref{ldloss} and $\lambda$ in Eq. \ref{total_loss} are set to 1 and 0.5, respectively. The raw data is randomly cropped into 224x224 pixel images with batches of 16, which are used as inputs for the fusion network. 

\subsection{Ablation Study}

\subsubsection{Ablation Study on Language-driven Fusion Loss}
As shown in Table \ref{abt_ldl}, the language-driven fusion loss (LDL) can comprehensively enhance the performance of infrared-visible image fusion. Without LDL, all metrics show a significant decline in their scores. Fig. \ref{abf_ld_obj}(c) and (f) show some fused results obtained from the method with and without LDL, respectively. We can see that the results obtained without LDL exhibit reduced performance in terms of the overall visual perception, clarity of details (see the orange boxes) and object saliency (see the yellow boxes). 

\begin{table}[t]
	\centering
	\renewcommand{\arraystretch}{0.9}
	\caption{Comparison of object detection performance based on the fused images with and without ``salient objects (SO)'' on M3FD (M1: mAP@0.5, M2: mAP@0.5:0.95).}
	\setlength{\tabcolsep}{1.5pt}
	\begin{tabular}{ccccccccc}
		\hline
		& & People & Car & Bus & Motorcycle & Lamp & Truck & All \\
		\hline
		\multirow{2}{*}{W/O SO} 
		& M1 & 0.760 & 0.885 & 0.876 & 0.689 & 0.663 & 0.795 & 0.778 \\
		& M2 & 0.450 & 0.643 & 0.709 & 0.347 & 0.389 & \textbf{0.586} & 0.520 \\
		\hline
		\multirow{2}{*}{LDFusion} 
		& M1 & \textbf{0.772} & \textbf{0.891} & \textbf{0.899} & \textbf{0.726} & \textbf{0.673} & \textbf{0.804} & \textbf{0.794} \\
		& M2 & \textbf{0.456} & \textbf{0.645} & \textbf{0.724} & \textbf{0.367} & \textbf{0.396} & 0.585 & \textbf{0.529} \\
		\hline
	\end{tabular}
	\label{abt_obj}
\end{table}

\begin{figure}[t]
	\centering
	\includegraphics[width=8cm]{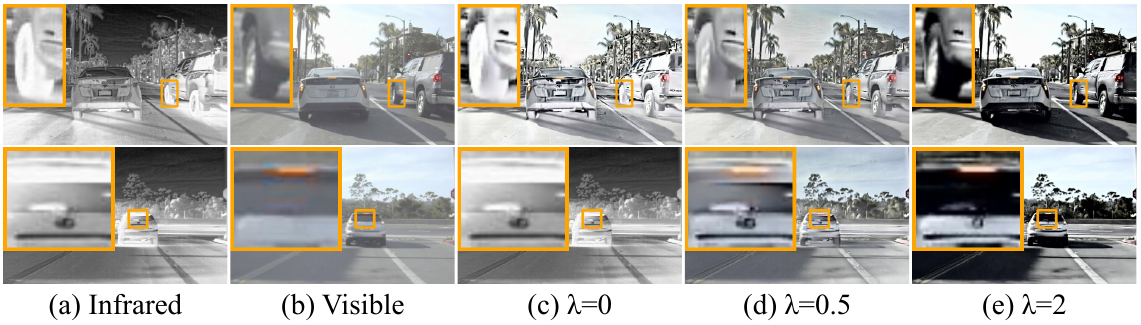} 
	\caption{Some fusion results with different $\lambda$ in language-driven fusion loss.}
	\label{abf_regulation}
\end{figure}

\begin{figure}[t]
	\centering
	\includegraphics[width=8cm]{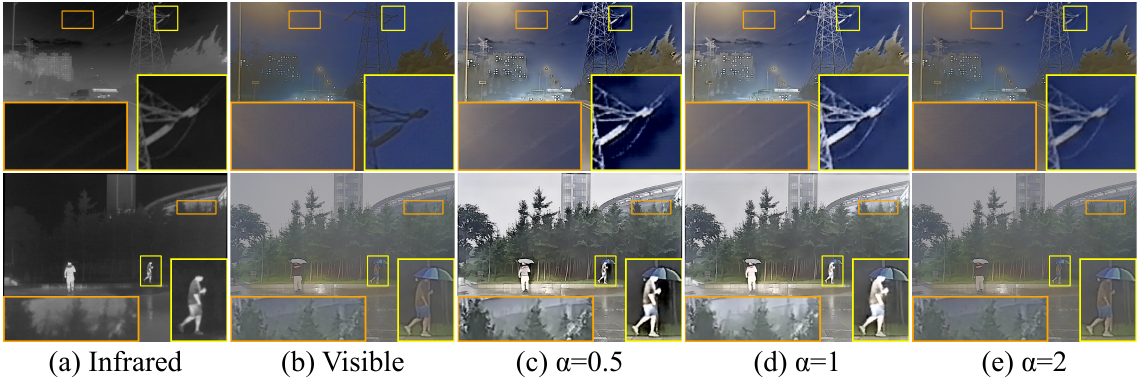} 
	\caption{Some fusion results with different $\alpha$ in total loss.}
	\label{abf_ldl}
\end{figure}

\begin{figure*}[!t]
	\centering
	\includegraphics[width=18cm]{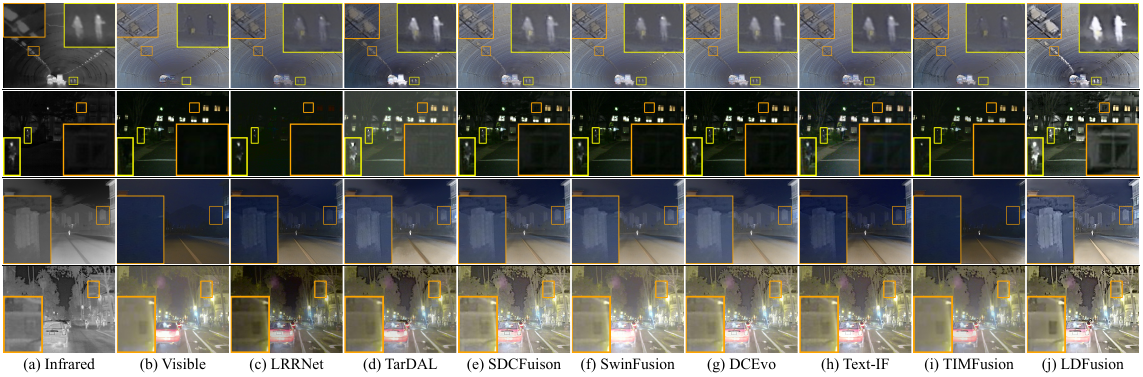} 
	\caption{Comparison of the fusion results from different methods. Text-IF uses its anti-degradation text for low-light.}
	\label{cp_results}
\end{figure*}

\subsubsection{Analysis on Language-expressed Fusion Objective}
We conduct three ablation experiments to analyze the roles of “salient objects”, “detailed background” and “clear image” in language-expressed fusion objective (LEFO), respectively.

\textbf{Ablation Study on “salient objects (SO)” in LEFO.} \indent Fig. \ref{abf_obj} shows that our LDFusion exhibits distinct advantages in the fusion of objects. Firstly, LDFusion improves the saliency of object in complex background (see the pedestrian crossing the street in first row). Secondly, LDFusion can significantly highlight hidden objects (see the person in orange boxes of second row). More importantly, incorporating ``salient objects'' in LEFO can also enhance the performance of downstream high-level vision task. For example, we utilize YOLOv8 \footnote{https://github.com/ultralytics/ultralytics} as the baseline model for object detection using the fused images. As shown in Table \ref{abt_obj}, including ``salient objects'' yields an obvious improvement on object detection task, indicating that it can effectively enhance the object semantic information through the fusion. It should be noted that using ``salient objects'' not only increases the visual saliency of the objects, but more importantly, it drives the fusion model to better combine or select the different information from source images for each individual object, so that they can be semantically salient in CLIP's cognition. Similarly, the same principle is also applicable to ``detailed background'' which will be demonstrated next.

\textbf{Ablation Study on “detailed background (DB)” in LEFO.} Without “detailed background”, critical details may not be sufficiently fused, leading to the loss of information. As illustrated in Fig. \ref{abf_ld_obj}(e) and (f), the fused images exhibit partial blurring of background content, such as the building structures (orange box in the first row), trees (orange box in the last row). In contrast, the method with DB can more effectively fuse the structures and details of the backgrounds from source images, and significantly improves the richness of details and overall quality.

\textbf{Ablation Study on “clear image (CI)” in LEFO.} The inclusion of “clear image” in LEFO benefits the fusion algorithm in two folds. Firstly, it can force the algorithm to find a more effective way to combine information from different source images to produce a composite image with clearer information. This may also involve excluding irrelevant information from the source images, especially for haze or smoke (for example, see the orange boxes in second row of Fig. \ref{abf_ld_obj}(d) and (f)). Second, it would enhance the clarity of the whole fused image as well, even in the low-light condition (see the first row).

\begin{table}[t]
	\centering
	\renewcommand{\arraystretch}{0.9}
	\caption{Comparison of fusion performance with different versions of prompts for source images.}
	\setlength{\tabcolsep}{2pt}
	\begin{tabular}{ccccccc}
		\hline
		Datasets & Methods & EN $\uparrow$ & CC $\uparrow$ & SD $\uparrow$ & EI $\uparrow$ & VIFF $\uparrow$  \\ \hline
		\multirow{3}{*}{TNO} 
		& swapped prompt & 6.976 & 0.734 & 38.484 & 42.214 & 0.521  \\ 
		& content prompt & 7.222 & 0.725 & 44.722 & 52.368 & 0.721  \\ 
		& LDFusion & \textbf{7.417}  & \textbf{0.741}  & \textbf{54.744} & \textbf{59.643} & \textbf{0.871}  \\ \hline
		\multirow{3}{*}{RoadScene} 
		& swapped prompt & 7.140 & 0.786 & 40.553 & 52.087 & 0.438  \\ 
		& content prompt & 7.409 & 0.782 & 49.097 & 75.666 & 0.577  \\ 
		& LDFusion & \textbf{7.529} & \textbf{0.792} & \textbf{52.511} & \textbf{76.861} & \textbf{0.622}  \\ \hline
	\end{tabular}
	\label{abt_source}
\end{table}

\begin{table*}[!t]
	\centering
	\renewcommand{\arraystretch}{0.9}
	\caption{Quantitative assessment of our LDFusion and compared methods on M3FD, TNO and RoadScene, with the best results highlighted in red and the second best in blue. (Higher values indicate better performance)}
	\setlength{\tabcolsep}{3pt}
	\begin{tabular}{c|ccccc|ccccc|ccccc}
		\hline
		\multirow{2}*{Methods} & \multicolumn{5}{c|}{M3FD Dataset} & \multicolumn{5}{c|}{TNO Dataset} & \multicolumn{5}{c}{RoadScene Dataset} \\
		\cline{2-16}
		& EN $\uparrow$ & CC $\uparrow$ & SD $\uparrow$ & EI $\uparrow$ & VIFF $\uparrow$ & EN $\uparrow$ & CC $\uparrow$ & SD $\uparrow$ & EI $\uparrow$ & VIFF $\uparrow$ & EN $\uparrow$ & CC $\uparrow$ & SD $\uparrow$ & EI $\uparrow$ & VIFF $\uparrow$ \\
		\hline
		LRRNet \cite{li2023lrrnet}         & 6.427 & \textcolor{red}{\textbf{0.771}} & 26.013 & 37.380 & 0.365 & 7.117 & 0.716 & 43.490 & 39.125 & 0.449 & 7.134 & \textcolor{red}{\textbf{0.793}} & 42.504 & 48.626 & 0.379 \\
		TarDAL \cite{liu2022target}       & \textcolor{blue}{\textbf{7.143}} & 0.709 & \textcolor{blue}{\textbf{42.673}} & 42.700 & 0.489 & 7.150 & 0.716 & 45.717 & 41.355 & 0.513 & 7.187 & 0.762 & 44.926 & 49.826 & 0.374 \\
		SDCFusion \cite{liu2024semantic}   & 6.949 & \textcolor{blue}{\textbf{0.738}} & 36.256 & \textcolor{blue}{\textbf{54.784}} & \textcolor{blue}{\textbf{0.596}} & 7.097 & \textcolor{blue}{\textbf{0.731}} & 40.423 & \textcolor{blue}{\textbf{50.020}} & \textcolor{blue}{\textbf{0.578}} & \textcolor{blue}{\textbf{7.314}} & 0.791 & 45.337 & \textcolor{blue}{\textbf{71.929}} & 0.545 \\
		SwinFusion \cite{ma2022swinfusion} & 6.781 & 0.726 & 35.359 & 47.594 & 0.488 & 6.908 & 0.722 & 39.735 & 42.252 & 0.451 & 6.877 & 0.791 & 44.224 & 46.776 & 0.441 \\
		DCEvo \cite{liu2025dcevo}    &6.841&0.720&35.579&48.163&0.435&6.951&0.715&39.389&41.381&0.400&7.184&0.775&44.663&57.763&0.443 \\
		Text-IF \cite{yi2024text}           & 6.860 & 0.718 & 35.478 & 52.915 & 0.530 & \textcolor{blue}{\textbf{7.226}} & 0.717 & \textcolor{blue}{\textbf{46.951}} & 48.159 & 0.550 & 7.302 & 0.776 & \textcolor{blue}{\textbf{50.171}} & 68.068 & \textcolor{blue}{\textbf{0.580}} \\
		TIMFusion \cite{liu2024task}        & 6.652 & 0.718 & 31.419 & 42.832 & 0.330 & 6.968 & 0.691 & 43.694 & 36.731 & 0.305 & 7.097 & 0.759 & 40.330 & 47.664 & 0.308 \\
		\hline
		LDFusion & \textcolor{red}{\textbf{7.341}} & 0.736 & \textcolor{red}{\textbf{46.667}} & \textcolor{red}{\textbf{65.443}} & \textcolor{red}{\textbf{0.792}} & \textcolor{red}{\textbf{7.417}} & \textcolor{red}{\textbf{0.741}} & \textcolor{red}{\textbf{54.744}} & \textcolor{red}{\textbf{59.643}} & \textcolor{red}{\textbf{0.871}} & \textcolor{red}{\textbf{7.529}} & \textcolor{blue}{\textbf{0.792}} & \textcolor{red}{\textbf{52.511}} & \textcolor{red}{\textbf{76.861}} & \textcolor{red}{\textbf{0.622}} \\
		\hline
	\end{tabular}
	\label{cpt_metrics}
\end{table*}

\subsubsection{Analysis of hyperparameters}
We analyze the influence of hyperparameters in our method, including $\lambda$ (the weight of direction regularization in Eq. (\ref{ldloss})) and $\alpha$ in Eq. (\ref{total_loss}).

\textbf{Analysis of $\lambda$.} In Eq. (\ref{ldloss}), the regularization term is important to improve the robustness and generalization of the fusion model. Without the expertly designed regularization, the trained fusion model may not perform very well in some cases, often resulting in fused results that are biased toward one of the modalities (for example, see Fig. \ref{abf_regulation}(c) where $\lambda$ = 0). In our implementation, we introduce a proper degree of regularization by setting $\lambda$ = 0.5, which can effectively improve the quality and robustness of fusion in practice (see the examples in Fig. \ref{abf_regulation}(d)). Besides, we also observed that an excessive regularization would in turn degrade the fusion performance, as illustrated in Fig. \ref{abf_regulation}(e) (where $\lambda$ = 2).

\textbf{Analysis of $\alpha$.} $\alpha$ is used to control the weight of the feature-fidelity loss in the total loss. A larger $\alpha$ places more emphasis on the feature fidelity, but might limit the fusion quality. However, a much smaller value of $\alpha$ could not ensure good information fidelity. As shown in Fig. \ref{abf_ldl}(c), when $\alpha$ is set as small as 0.5, obvious black halos are generated (the yellow boxes), decreasing the information fidelity. As $\alpha$ increases to 2, some details could be lost (see the orange boxes) and the fused information could lose clarity, resulting in a lower fusion quality. As a result, we use a moderate setting of $\alpha$ = 1, which achieves satisfactory results in practice.

\subsubsection{Ablation Study on Language Prompts of Source Images}
\label{ab_source}
To analyze the role of prompts for source images, we employ three distinct types of prompts for comparison: (1) the original prompts (as described in Sec. \ref{prompt_source}); (2) the swapped prompts (\textit{i.e.}, “an infrared image” for the visible image and “a visible gray image” for the infrared image); and (3) content-based prompts (\textit{i.e.}, the description of detailed content in images). The fusion results are summarized in Table \ref{abt_source}. 

We can see that the swapped version obviously degrades the fusion performance, due to their poor alignment with the infrared and visible source images in the CLIP space.  Although the content texts align well with the source images, they lead to slightly inferior generalization performance for the fusion compared with the original prompts as can be seen in Table \ref{abt_source}, where the results were obtained by the fusion network trained on M3FD but tested on TNO and RoadScene datasets. A similar conclusion can also be drawn when training on TNO or RoadScene and testing on the other dataset. This might be because the content texts tend to rely on the specific content in individual images, making them less robust for generalization across different datasets. In contrast, the original prompts, although fixed and not content-adaptive, can at least maintain sufficiently good alignment (as illustrated in Fig. \ref{clip_aware}) despite the content variation in source images. In summary, a practical insight is that maintaining alignment irrespective of varied content is more important than being content-adaptive.  

\subsection{Comparison with State-of-the-art Methods}
\subsubsection{Qualitative Comparison}
Some qualitative results of our LDFusion and compared methods are presented in Fig. \ref{cp_results}. As shown in Fig. \ref{cp_results}, our LDFusion significantly enhances the fusion quality for the objects. While SDCFusion, TarDAL, and TIMFusion utilize high-level tasks (\textit{e.g.}, object detection, semantic segmentation) to supervise the fusion process, they still can not achieve competitive fusion performance on objects when compared to our method (see the yellow boxes). Moreover, our method can obtain much clearer and sharper details. In particular, the close-up view on the house (the orange box in third and fourth rows) shows that our method perfectly fuses the details and structures of the house from different source images and presents a vivid appearance of it, showing the impressive powerful of the language-driven fusion loss. 

\subsubsection{Quantitative Comparison}
Table \ref{cpt_metrics} summarizes the quantitative results of LDFusion and all compared SOTA methods on various evaluation metrics. Our method achieves the highest average scores on almost all metrics on the datasets M3FD, TNO and RoadScene. It should be noted that our test model was only trained on some portions of the M3FD dataset (with the other portion serving as the test dataset). In this case, we can see that the model still performs very well when tested on the TNO and RoadScene datasets, demonstrating the good generalization capability of the proposed method.

\subsubsection{Performance on High-level Task}

\begin{table}[t]
	\centering
	\renewcommand{\arraystretch}{0.9}
	\caption{Quantitative comparison of object detection performance on M3FD, with the best results highlighted in bold.}
	\begin{tabular}{ccccc}
		\hline
		Methods & \scalebox{0.9}{Precision} & \scalebox{0.9}{Recall} & \scalebox{0.9}{mAP@0.5} & \scalebox{0.9}{mAP@0.5:0.95} \\
		\hline
		LRRNet      & 0.826 & 0.709 & 0.773  & 0.517 \\
		TarDAL      & 0.834 & 0.712 & 0.779  & 0.520 \\
		SDCFusion   & 0.832 & 0.706 & 0.779  & 0.525 \\
		SwinFusion  & 0.833 & 0.701 & 0.775  & 0.518 \\
		DCEvo       & 0.846 & 0.713 & 0.784  & 0.520 \\
		Text-IF     & 0.838 & 0.704 & 0.782  & 0.523 \\
		TIMFusion   & 0.863 & 0.695 & 0.779  & 0.519 \\
		\hline
		LDFusion    & \textbf{0.865} & \textbf{0.716} & \textbf{0.794} & \textbf{0.529} \\
		\hline
	\end{tabular}
	\label{cpt_detect}
\end{table}

To verify the performance of our method, we conduct object detection experiments with the fusion results of different methods. We employ YOLOv8 as the object detection baseline, and train separate instances of YOLOv8 using the fused images of each fusion method. The quantitative detection results are presented in Table \ref{cpt_detect}. Our fusion method achieves the best object detection performance with the highest values in precision, recall, mAP@0.5, and mAP@0.5:0.95, which implies that our method can integrate more semantic information from source images. Although TIMFusion and TarDAL employ the high-level task of object detection to supervise the fusion process, their detection performance is not as good as ours. 

\section{Conclusion}
In this paper, we use natural language to express the whole IVIF objective thereby avoiding complex and explicit mathematical modeling in fusion loss functions. To achieve this goal, we construct an embedded language-driven fusion model based on the CLIP text encoder, which defines the expected fusion directions within CLIP embedding space. We then propose a language-driven loss to align the actual image fusion with the embedded model through supervised training, thereby generating fused images consistent with the language-expressed fusion objective. Furthermore, a novel patch-filtering based training approach is proposed to solve the challenge of textual artifact removal for the fusion method. Extensive experiments demonstrate that our method can significantly improve the quality of IVIF, and we have provided a detailed analysis and ablation study helping to better understand the effectiveness and underlying mechanisms of the proposed method. 

\clearpage 

\begin{acks}
This work was supported by grants from the National Natural Science Foundation of China (62471045, 62071036, 62173040).
\end{acks}

\bibliographystyle{ACM-Reference-Format}
\bibliography{fusion.bib}

\end{document}